%% file: NIPS_arXiv_version.tex
\documentclass{article}

\usepackage[preprint, nonatbib]{nips_2018}

\usepackage[utf8]{inputenc} 
\usepackage[T1]{fontenc}    
\usepackage{hyperref}       
\usepackage{url}            
\usepackage{booktabs}       
\usepackage{amsfonts}       
\usepackage{amsmath}
\usepackage{nicefrac}       
\usepackage{microtype}      
\usepackage{color}
\usepackage{graphicx}
\usepackage{subcaption}
\usepackage{xspace}

\input{tikzstuff.tex}

\def\KL{\mathcal{D}_\text{KL}}

\title{Gaussian mixture models with Wasserstein distance}

\author{
  Benoit Gaujac\\
  University College London \\
    \And
  Ilya Feige \\
  ASI Data Science \\
    \And
  David Barber \\
  University College London \\
  Alan Turing Institute \\
}

\begin{document}

\maketitle

\begin{abstract}
Generative models with both discrete and continuous latent variables are highly motivated by the structure of many real-world data sets. They present, however, subtleties in training often manifesting in the discrete latent being under leveraged. In this paper, we show that such models are more amenable to training when using the Optimal Transport framework of Wasserstein Autoencoders. We find our discrete latent variable to be fully leveraged by the model when trained, without any modifications to the objective function or significant fine tuning. Our model generates comparable samples to other approaches while using relatively simple neural networks, since the discrete latent variable carries much of the descriptive burden. Furthermore, the discrete latent provides significant control over generation.
\end{abstract}

\section{Introduction}
\label{sec:intro}

Unsupervised learning using generative latent variable models provides a powerful and general approach to learning the underlying, low-dimensional structure from large, unlabeled datasets. Perhaps the two most common techniques for training such models are Variational Autoencoders (VAEs) \cite{kingma2014stochastic, rezende2014stochastic}, and Generative Adversarial Networks (GANs) \cite{GANs}. Both have advantages and disadvantages. VAEs provide a meaningful lower bound of the log likelihood that is stable under training as well as an encoding distribution from the data to the latent space. However, they generate blurry samples due to the VAE objective being unable to handle deterministic decoders and tractability requiring simple priors \cite{ELBO_surgery}. On the other hand, GANs naturally enable deterministic generative models with sharply defined samples, but their training procedure is much less stable \cite{ArjovskyB17}.

A relatively new approach to training generative models has emerged based on minimizing the Optimal Transport (OT) distance \cite{villani2008optimal} between the generative model distribution and the data distribution. The OT approach provides a general framework for training generative models, which promises some of the best of both GANs and VAEs. Though interesting first results have been given in \cite{WGAN, WAE_followup, WAE}, the OT approach to generative modelling is still nascent.

Our contributions are twofold: we seek both to improve generative modelling capabilities with discrete and continuous latent variables, and to show that training generative models with OT in particular cases can be significantly more effective than the traditional VAE approach.

Discrete latent-variable models are critical to the endeavor of unsupervised learning because of the ubiquity of discreteness in the natural world, and hence in the datasets that describe it. However, they are harder to train than their continuous counterparts.
This has been tackled in a number of ways (e.g., directly mitigating high-variance discrete samples \cite{FastInferRepeat, Lawson2017LearningHA}, parametrizing discrete distributions using continuous ones \cite{GumbelSoftmax, Concrete, NeuralDiscrete}, deliberate model design leveraging conjugacy \cite{SVAE}).

However, even in the simple case where the number of mixtures is small enough that monte-carlo sampling from the discrete latent is avoidable, training can still be problematic. For example, in \cite{HackGMVAE} a Gaussian-mixture latent-variable model (GM-LVM) was studied, and the authors were unable to train their model on MNIST using variational inference without substantially modifying the VAE objective. What appears to happen is that the model quickly learns to ``hack'' the VAE objective function by collapsing the discrete latent variational distribution. This problem only occurs in the unsupervised setting, as \cite{kingma2014semi} are able to learn the discrete latent in the semi-supervised version of the same problem once they have labeled samples for the discrete latent to latch onto. This is discussed in more detail in Section~\ref{sec:VAE_difficulty}.

The OT approach to training generative models (in particular the Wasserstein distance, discussed in Section~\ref{sec:OT_easy}) induces a weaker topology on the space of distributions, enabling easier convergence of distributions than in the case of VAEs \cite{Bousquetetal17}. Thus, one might conjecture that the OT approach would enable easier training of GM-LVMs than the VAE approach. We provide evidence that this is indeed the case, showing that GM-LVMs can be trained in the unsupervised setting on MNIST, and motivating further the value of OT in generative modelling.



\section{Gaussian-Mixture Wasserstein Autoencoders}
\label{sec:GM-WAEs}

We consider a hierarchical generative model $p_G$ with two layers of latent variables, the highest one being discrete. Explicitly, if we denote the discrete latent $k$ with density $p_D$ ($D$ for discrete), and the continuous latent $z$ with density $p_C$ ($C$ for continuous), the generative model is given by:
\begin{equation}
    \label{eq:GM-LVM}
    p_G(x) = \sum_{k}\int dz \; p_G(x | z) \, p_C(z|k) \, p_D(k)
\end{equation}


In this work, we choose a categorical distribution $p_D = \text{Cat}(K)$ and a continuous distribution $p_C(z|k) = \mathcal{N}(\mu_k^0;\Sigma_k^0)$.
We refer to this GM-LVM as a GM-VAE when it is trained as a VAE \cite{kingma2014stochastic, rezende2014stochastic} or GM-WAE when trained as a Wasserstein Autoencoder \cite{WAE} (discussed in Section~\ref{sec:OT_easy}).

A prior structured as such is motivated when the data is expected to be composed of $K$ different classes of objects.
For example in images, while the data lie in a continuous and low dimensional manifold, each object that appears would be described by a separate mode within this manifold.

\subsection{The difficulty of training GM-VAEs}
\label{sec:VAE_difficulty}

Training GM-LVMs in the traditional VAE framework (GM-VAEs) would involve maximizing the evidence lower bound (ELBO) averaged over the data. 
Such models are empirically hard to train \cite{HackGMVAE}. This is likely due to the fact that the discrete latent variational distribution learns on a completely different scale from the generative distribution, due to its simplicity. Consequently, the discrete latent variational distribution tends to instantly learn some unbalanced structure where its classes are meaningless in order to accommodate the untrained generative distribution. The generative model then learns around that structure, galvanizing the meaningless discrete distribution early in training.

We choose a variational distribution $q(z,k|x) = q_C(z|k,x)q_D(k|x)$ to mirror the prior in Eq.~\eqref{eq:GM-LVM}. With this, the ELBO can be written as follows:
\begin{align}
    \text{ELBO} 
    &= \mathbb{E}_{q_D(k|x)} \Big[ \mathbb{E}_{q_C(z|k,x)}\big[\log p_G(x|z)\big] - \KL\big( q_C(z|k,x) || p_C(z|k) \big) \Big]
    \label{eq:ELBO_breakdown}\\\notag& 
    -\KL\big( q_D(k|x) || p_D(k) \big)
\end{align}
The separation of the discrete $\KL$ term is possible because the ELBO is evaluated separately for each data point, an attribute of VAEs that is not shared with WAEs.

\begin{figure}
\begin{subfigure}{0.33\textwidth}
    \centering\includegraphics[width=\textwidth]{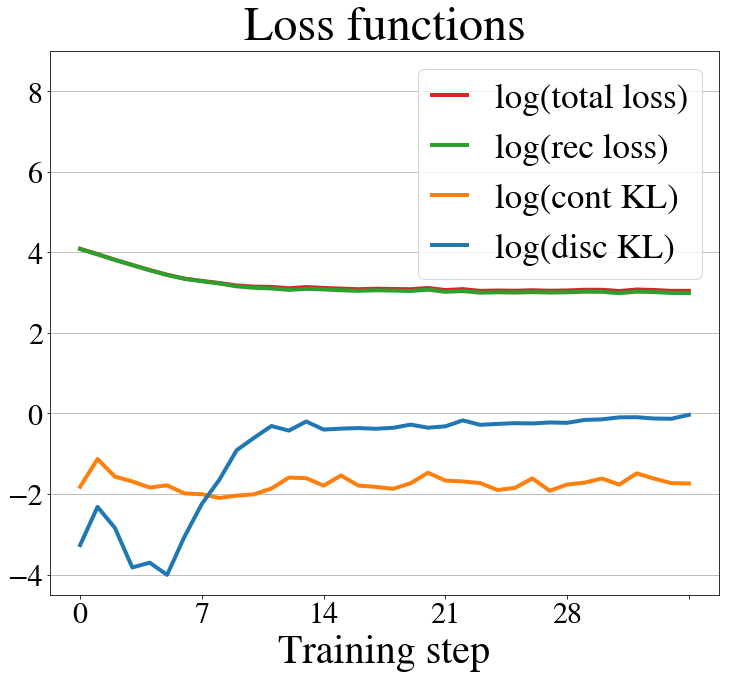}
    \caption{}
    \label{fig:early_VAE_loss}
\end{subfigure}
\hspace{.02\textwidth}
\begin{subfigure}{0.32\textwidth}
    \centering\includegraphics[width=\textwidth]{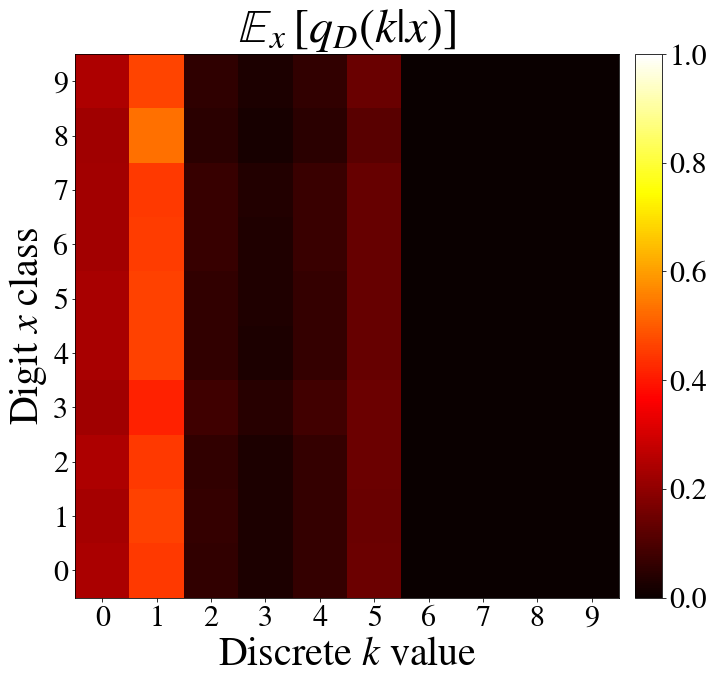}
    \caption{}
    \label{fig:early_VAE_probs}    
\end{subfigure}
\hspace{.01\textwidth}
\begin{subfigure}{0.3\textwidth}
    \centering\includegraphics[width=\textwidth]{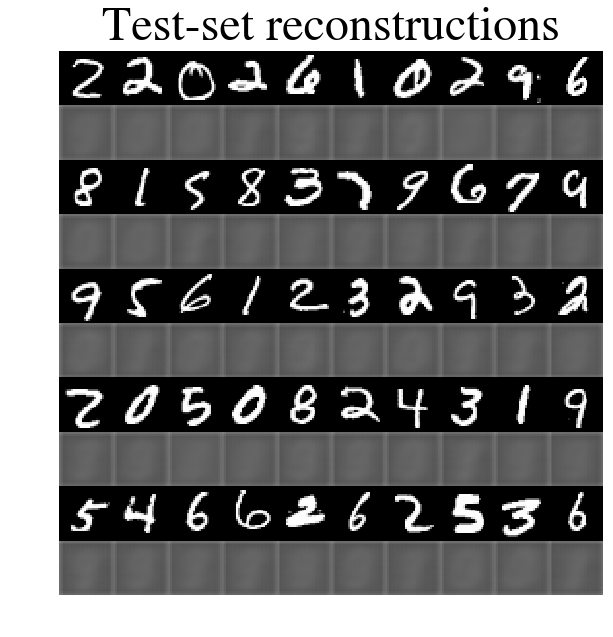}
    \caption{}
    \label{fig:early_VAE_recons}    
\end{subfigure} \\
\begin{subfigure}{0.33\textwidth}
    \centering\includegraphics[width=\textwidth]{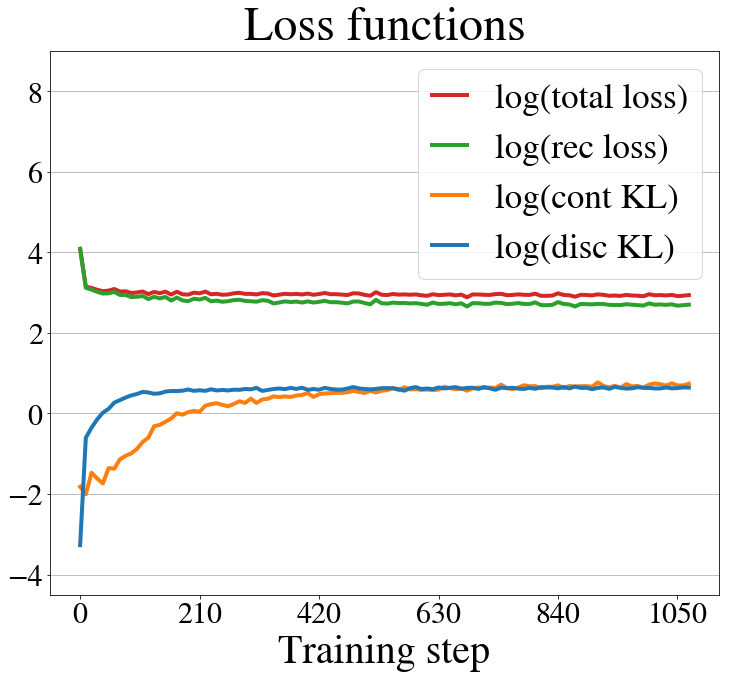}
    \caption{}
    \label{fig:mid_VAE_loss}
\end{subfigure}
\hspace{.02\textwidth}
\begin{subfigure}{0.32\textwidth}
    \centering\includegraphics[width=\textwidth]{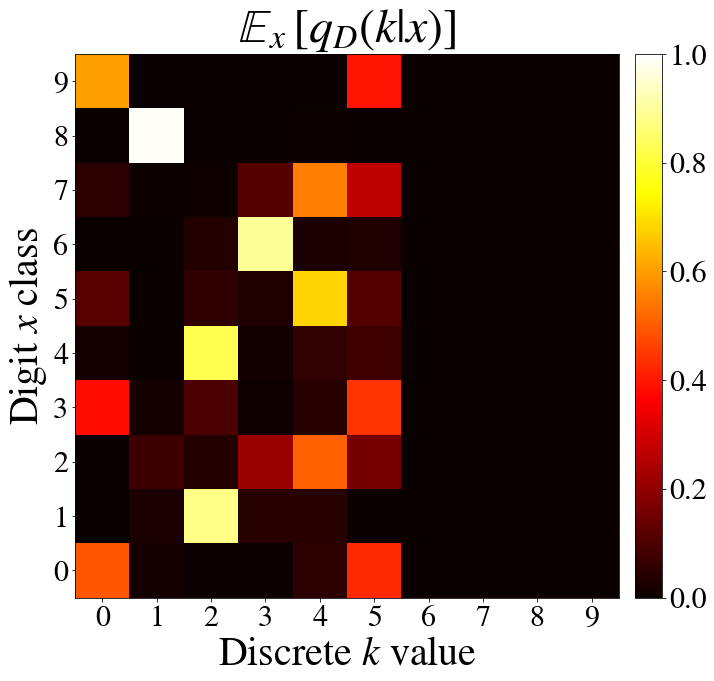}
    \caption{}
    \label{fig:mid_VAE_probs}    
\end{subfigure}
\hspace{.01\textwidth}
\begin{subfigure}{0.3\textwidth}
    \centering\includegraphics[width=\textwidth]{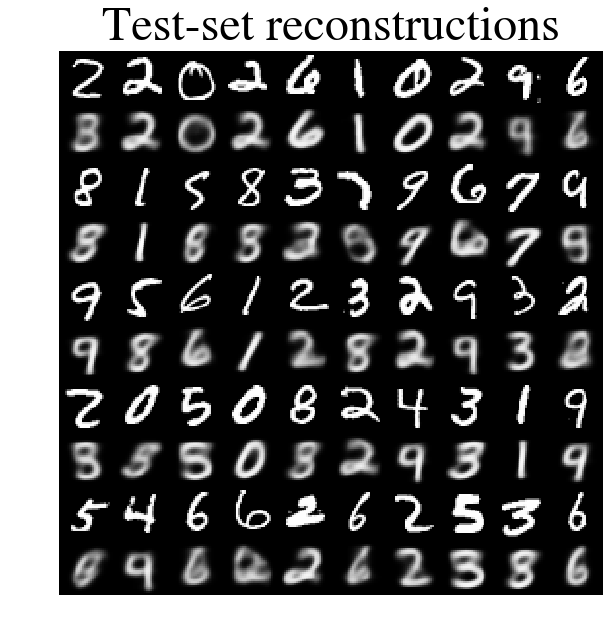}
    \caption{}
    \label{fig:mid_VAE_recons}    
\end{subfigure}
\caption{(a), (b), (c) show a snapshot of the GM-VAE after the first 35 training steps. The loss curves are shown in (a), the discrete variational distribution in (b) with rows $\ell$ representing $\mathbb{E}_{\{x|\text{label}(x)=\ell\}} q_D(k|x)$, and reconstructions from the GM-VAE shown in (c). Similarly, (d), (e), (f) show a snapshot of the same GM-VAE after approximately 1000 training steps.}
\label{fig:VAE_local_minima}
\end{figure}

Both the first and the second lines in Eq.~\eqref{eq:ELBO_breakdown} depend on $q_D(k|x)$. However, the term on the second line is much smaller than the above terms (it is bounded by $\log K$ for uniform $p_D$ over $K$ classes, whereas the other terms are unbounded from above; though we will initialize the modes of $q_C$ to match those of the priors making the continuous KL term initially small as well). As a consequence, $q_D(k|x)$ will immediately shut off multiple $k$ values (i.e., $q_D(k|x) = 0\; \forall x$ and multiple $k$s) with large reconstruction loss, $\mathbb{E}_{q_C(z|k,x)}\big[\log p_G(x|z)\big]$. This is shown in the top row of Figure~\ref{fig:VAE_local_minima} where within the first 20 training steps the reconstruction loss has substantially decreased (Figure~\ref{fig:early_VAE_loss}) by simply shutting off 4 values of $k$ in $q_D(k|x)$ (Figure~\ref{fig:early_VAE_probs}). In Figure~\ref{fig:early_VAE_loss} it can be seen that the discrete KL term increases drastically and concurrently with the decrease in reconstruction loss. However, this drastic increase in the discrete KL term is negligible since the term is multiple orders of magnitude smaller than the reconstruction term in the ELBO. All of this takes place in the first few training iterations; well before the generative model has learned to use its continuous latent (see Figure~\ref{fig:early_VAE_recons}).

Subsequently, on a slower timescale, the generative model starts to learn to reconstruct from its continuous latent, causing $q_C(z|k,x)$ to shift away from its prior toward a more-useful distribution to the generative model. This can be seen in Figure~\ref{fig:mid_VAE_loss} through the growth of the continuous KL curve which takes place concurrently to the downturn of the reconstruction loss term. Figure~\ref{fig:mid_VAE_recons} shows that after this transition, which in the case shown takes place within the first thousand training steps, the reconstructions from the model start to look more like MNIST digits.

While the generative model learns to use the continuous latent, the discrete distribution $q_D(k|x)$ never revives the $k$ values that it shut off. This is because the generative model would not know how to use the $z\sim q_C(z|k,x)$ values for those $k$s, implying a significant penalty in the reconstruction term of the ELBO. This is evidenced in Figure~\ref{fig:mid_VAE_loss} by the discrete KL staying flat, and in Figure~\ref{fig:mid_VAE_probs} where the columns corresponding to the shut off $k$ values never repopulate.

We have discussed the difficulty in the convergence of GM-VAEs in detail using our specific implementation designed to mirror the GM-WAE of Section~\ref{sec:OT_easy} (we also considered many other variants which performed similarly). Though the root cause of this difficulty has not been ascertained in generality, we expect it to be in part due to the per-data-point nature of the ELBO objective, in particular, the impact of KL divergence term on learning the variational distribution.

\subsection{Optimal Transport facilitates training of GM-LVMs}
\label{sec:OT_easy}

The difficulty associated with training GM-VAEs may be interpreted as a problem of restricted convergence of a sequence of distributions, where the sequence is indexed by the training steps. If that were so, an objective function that induces a weaker topology might help GM-LVMs converge to a distribution that non-trivially uses its discrete latent variable. Hence, we are motivated to consider approaching the training of such models using the OT framework, and in particular the Wasserstein distance objective function. We do not attempt to make this connection rigorous. However, this was indeed the motivation for the authors to apply OT techniques to this problem.

Following the OT approach of \cite{WAE}, we would like to minimize the 2-Wasserstein distance between the underlying data distribution (from which we have samples) and our GM-LVM, namely:
\begin{equation}
W^{\dagger}_2\big(p_\text{data},p_G\big) =
\underset{\mathbb{E}_{p_\text{data}(x)}[ q(z,k|x)] = p_C(z|k) p_D(k)}
{\inf_{q(z,k|x)\in{\mathcal{P}_{\mathcal{Z}\times\mathcal{K}}}}}
\mathbb{E}_{p_\text{data}(x)}
\mathbb{E}_{q(z,k|x)}
\mathbb{E}_{p_G(y|z)} \big[||x-y||_2^{\;2}\big]
\label{eq:obj}
\end{equation}
where $\mathcal{P}_{\mathcal{Z}\times\mathcal{K}}$ is the set of all joint distributions over $z$ and $k$, such that $q(z,k|x) = q_C(z|k,x)q_D(k|x)$ with $q_C$ and $q_D$ parametrized below. Any parametrization of $q(z,k|x)$ 
reduces the search space of the infimum, so $W^{\dagger}_2$ is in fact an upper bound on the true 2-Wasserstein distance $W_2$. We choose to model the ``variational'' distribution $q(z,k|x)$ deliberately to mirror the structure of the prior, which differs from \cite{AAE} who assume conditional independence between $z|x$ and $k|x$.

Since the constrained infimum is intractable, a relaxed version of the 2-Wasserstein distance is introduced as follows:
\begin{align}
\widetilde W^{\dagger}_2\big(p_\text{data},p_G\big) &=
\inf_{q(z,k|x)\in{\mathcal{P}_{\mathcal{Z}\times\mathcal{K}}}}
\mathbb{E}_{p_\text{data}(x)}
\mathbb{E}_{q(z,k|x)}
\mathbb{E}_{p_G(y|z)} \big[||x-y||_2^{\;2}\big]
\label{eq:obj_relaxed} \\&+
\lambda\,\mathcal{D}\bigg(
\mathbb{E}_{p_\text{data}(x)}\big[q(z,k|x)\big]
\Big|\Big|
p_C(z|k) p_D(k) \bigg)
\notag
\end{align}
which is equivalent to the original distance in the limit where $\lambda\to\infty$. This equivalence requires only that $\mathcal{D}$ be a divergence. We use the Maximum Mean Discrepancy (MMD), as it is a distance on the space of densities \cite{MMD}, and thus shares the properties of divergence functions, and it has an unbiased U-estimator \cite{WAE}. Explicitly if $K$ is a reproducing positive-definite kernel and is characteristic, then the MMD associated to $K$ is given by 
\begin{equation}
    \text{MMD}\big(q || p\big) = \mathbb{E}_{z_1,z_2\sim q} [K(z_1,z_2)] 
    + \mathbb{E}_{z_1,z_2\sim p}[K(z_1,z_2)] 
    - 2\mathbb{E}_{z_1\sim q, z_2\sim p}[K(z_1,z_2)]
\end{equation}
As in \cite{WAE}, we used a mixture of inverse multiquadratic (IMQ) kernels with different bandwidth $C^i$:
\begin{equation}
    K(z_1,z_2) = \sum_i K_{C^i}(z_1,z_2),
    \quad
    \text{where }\, \forall{i}, \; K_{C^i}(z_1,z_2)=\frac{C^i}{C^i+||z_1-z_2||_2^{\;2}}
\end{equation}
and we take $C^i\in[0.1, 0.2, 0.5, 1.0, 2.0, 5.0, 10.0]$ without tuning as was done in \cite{WAE_followup}. IMQ kernels have fatter tails than the classic radial basis function kernels, proving more useful early in training when the encoder has not yet learned to match the aggregated posterior with the prior. The choice of bandwidth for the kernel can be fickle, so we take a mixture of bandwidths reducing our sensitivity on any one choice (see \cite{gen_MMD,kernel_choice,moments_matching}).

The objective function for minimization is fully specified: the 2-Wasserstein distance $\widetilde W^{\dagger}_2(p_\text{data},p_G)$. We now turn to specifying our parametrization of the model.

As mentioned briefly in Section~\ref{sec:intro}, VAEs have the disadvantage that deterministic generative models cannot be used. As can be seen in Eq.~\eqref{eq:obj_relaxed}, this is not the case for with the Wasserstein distance. Thus we parametrize the generative density $p_G(x|z)$ as a deterministic distribution $x|z = g_\theta(z)$ where $g_\theta$ is a function from the latent to the data space specified by a deep neural network. With the model parameterized as a neural network, we would simply minimize the objective function using stochastic gradient descent with automatic differentiation.

However, even with $p^\theta_G(x|z)$ parametrized, there is still an infimum and multiple expectations with respect to the variational distribution $q(z,k|x)$ in the definition of the objective function. The infimum is approximated by parameterizing $q^\phi(z,k|x) = q_C^{\phi_C}(z|k,x) q_D^{\phi_D}(k|x)$ and performing gradient descent to minimize the objective function with respect to $\{\phi_D, \phi_C\}$ . This is an approximation since there is no guarantee that the infimum is achieved for any given parametric distribution.

To mirror the prior, we parameterize $q_C^{\phi_C}(z|k,x)$ as a Gaussian with diagonal covariance for each different $k$. This parameterization allows for the use of the reparameterization trick \cite{kingma2014stochastic, rezende2014stochastic} in order to compute gradients with lower variance. To avoid the problem of back propagating through discrete variables, where the reparameterization trick is not possible, we choose to compute the expectation over the discrete distribution $q_D^{\phi_D}(k|x)$ exactly, as 
MNIST modeling only requires a small number of classes making this expectation tractable
(this assumption can be relaxed via \cite{brooks2011handbook, GumbelSoftmax, Concrete}).

As previously mentioned, the weakness of the induced topology on the space of distributions may be the underlying reason to expect optimizing the Wasserstein distance to overcome the issues with training VAEs presented in Section~\ref{sec:VAE_difficulty}. With the model and objective function in hand, a less-abstract argument can be made in support of this claim.

Recall from Section~\ref{sec:VAE_difficulty} that the problem with the GM-VAE was that the objective function demands the various distributions be optimized at the individual data-point level. For example, the $\KL\big( q_D(k|x) || p_D(k) \big)$ term in Eq.~\eqref{eq:ELBO_breakdown} breaks off completely and becomes irrelevant due to its size. This causes the $q_D(k|x)$ distribution to shut off $k$ values early, which becomes galvanized as the generative model learns. 

However, in posing the problem in terms of the most efficient way to move one distribution $p_G$ onto another $p_\text{data}$, via the latent distribution $q(z,k|x)$, the Wasserstein distance never demands the similarity of two distributions at the individual data-point level. Indeed, the $\mathbb{E}_{p_\text{data}}$ in Eq.~\eqref{eq:obj_relaxed} is inside both the infimum and the divergence $\mathcal{D}$. We expect that ``aggregating'' the posterior as such will allow $q(z,k|x)$ (in particular, $q_D(k|x)$) the flexibility to learn data-point specific information while still matching the prior on aggregate. Indeed, it is also found in \cite{AAE} that when using an adversarial game to minimize the distance between an aggregated posterior and the prior, unsupervised training on MNIST with a discrete-continuous latent-variable model is highly successful.

\section{Results}
\label{sec:results}

In this work we primarily seek to show the potential of GM-LVMs and how OT techniques are effective at enabling their training. Thus, we use relatively simple neural network architectures and train on MNIST.

We use a mixture of Gaussians for the prior, with 10 mixtures to represent the 10 digits in MNIST and a non-informative uniform prior over these mixtures. Namely, for each $k\in\{0,\ldots,9\}$:
\begin{equation}
    p_D(k) = \frac{1}{10}
    , \qquad
    p_C(z|k) = \mathcal{N}(\mu_k^0;\sigma_k^0 I)
\end{equation}
where the $\mu_k^0$ are chosen to be mutually equidistant. We found that choosing $\text{dim}(z) = 10$ worked well. 
For each $k$, $\sigma_k^0 = \sigma^0$ is chosen identically in order to admit $\approx 5\%$ overlap between the 10 different Gaussian modes of the prior (i.e., the distance between any pair of means $\mu_k^0$ is $4\sigma^0$).

For the variational distribution, we take $q(z,k|x) = q_C(z|k,x)q_D(k|x)$ with
\begin{equation}
    q_D(k|x) = \pi_k(x) 
    ,\qquad
    q_C(z|k,x) =  \mathcal{N} \big(\mu_k(x); \text{diag}(\sigma_k(x))\big)
\end{equation}
where each component is parametrized by a neural network. For $\pi(x)$ a 2-layer DCGAN-style network \cite{DCGAN} is used with largest convolution layer composed of 32 filters. The Guassian networks $\mu_k(x), \sigma_k(x)$ are taken to be 16-unit single-hidden-layer dense networks. Finally, for the generative model, we take $p_G^\theta(x|z)$ to be deterministic with $x|z = g_\theta(z)$, using a 3-layer DCGAN-style network with smallest deconvolution layer composed of 128 filters.

The discrete-continuous structure in the variational distribution allows for relatively simple networks for the Gaussian latents, as each pair $(\mu_k, \sigma_k)$ must only capture the structure of the data within the $k$-th mode. Thus by splitting the expressiveness of model between the continuous and the discrete latents, we can achieve good performance using simple encoding networks. For comparison, in \cite{WAE} the generative model is a 4-layer DCGAN-style network with 1028 filters in its largest convolution.

We used batch normalisation \cite{BatchNorm}, ReLU activation functions after each hidden layer in both the encoder and the decoder network, as well as Adam for optimization \cite{ADAM} with a learning rate of $0.0008$. As in \cite{WAE}, we find that $\lambda=10$ works well. The $(\mu_k,\sigma_k)$ networks are pretrained to match the prior moments, which accelerates training and improves stability (this was also done for GM-VAE in Section~\ref{sec:VAE_difficulty}).

\subsection{Reconstructions and samples}
\label{sec:recon_samples}

\begin{figure}
\begin{center}
\begin{subfigure}{0.32\textwidth}
    \centering\includegraphics[width=\textwidth]{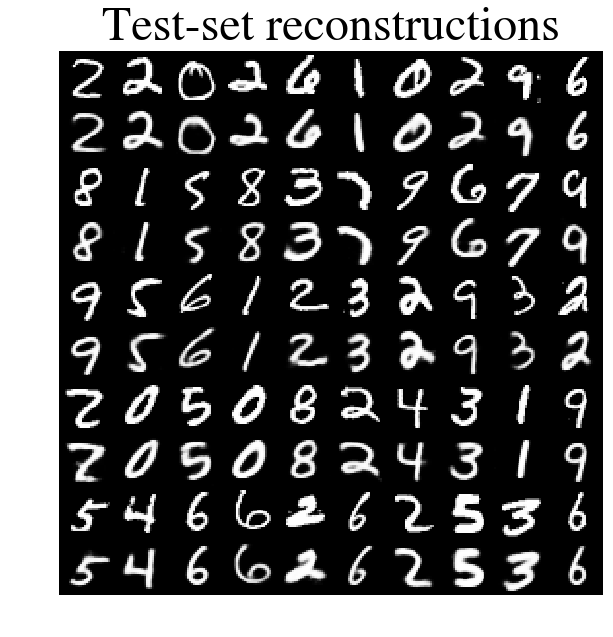}
    \caption{}
    \label{fig:test_recon}    
\end{subfigure}
\begin{subfigure}{0.32\textwidth}
    \centering\includegraphics[width=\textwidth]{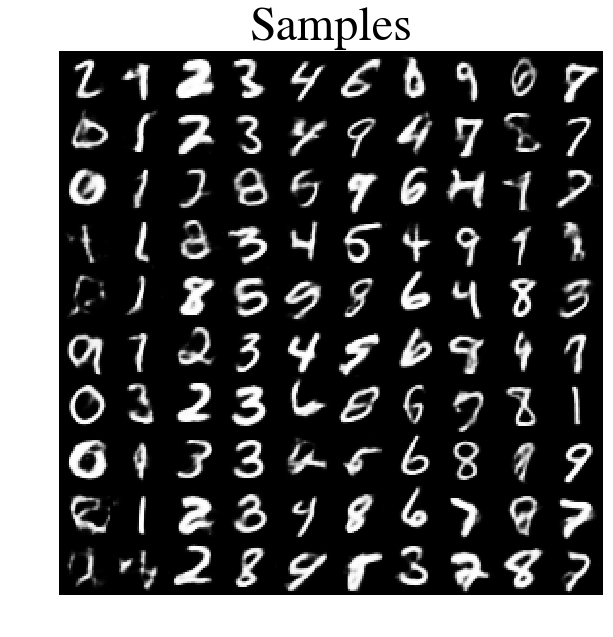}
    \caption{}
    \label{fig:generation_unfettered}
\end{subfigure}
\begin{subfigure}{0.32\textwidth}
    \centering\includegraphics[width=\textwidth]{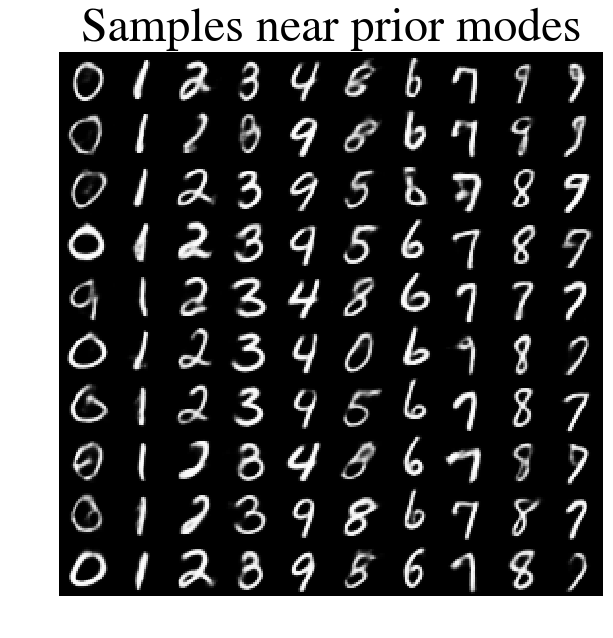}
    \caption{}
    \label{fig:generation_tight}
\end{subfigure}
\caption{Shown in (a) are reconstructions of the held-out test data from the inferred latent variables, $k \sim q_D(k|x)$ and $z \sim q_C(z | k, x)$. The first, third, fifth, etc, rows are the raw data, and the rows below show the corresponding reconstructions. Digit samples $x \sim p_G(x|z)p_C(z|k)$ for each discrete latent variable $k$ are shown in (b) as well as those samples closer to each mode of the prior are shown in (c). Since the discrete prior $p_D(k)$ is uniform, samples in (b) are representative of generation from the prior, only with columns ordered by discrete latent value. To get the samples in (c) close to each mode of the prior, we use $z$ values sampled from Gaussian distributions identical to $p_C(z|k)$, except with standard deviation reduced by $\frac12$.}
\label{fig:reconstructions}
\end{center}
\end{figure}

Our implementation of GM-WAE is able to reconstruct MNIST digits from its latent variables very well. In Figure~\ref{fig:test_recon} example data points from the held-out test set are shown on the odd rows, with their reconstructions on the respective rows below. 
The encoding of the input points is a two step process, first determining in which mode to encode the input via the discrete latent, and then drawing the continuous encoding from the corresponding mode.

Samples from the GM-WAE are shown in Figure~\ref{fig:generation_unfettered} and \ref{fig:generation_tight}. Since the discrete prior $p_D(k)$ is uniform, we can sample evenly across the $k$s in order from $0$ through $9$, while still displaying representative samples from $p(z,k) = p_C(z|k) p_D(k)$. Again, this shows how the GM-WAE learns to leverage the structure of the prior, whereas the GM-VAE results in the collapse of the several modes of the prior.

\begin{figure}
\begin{center}
\begin{subfigure}{0.49\textwidth}
    \centering\includegraphics[width=\textwidth]{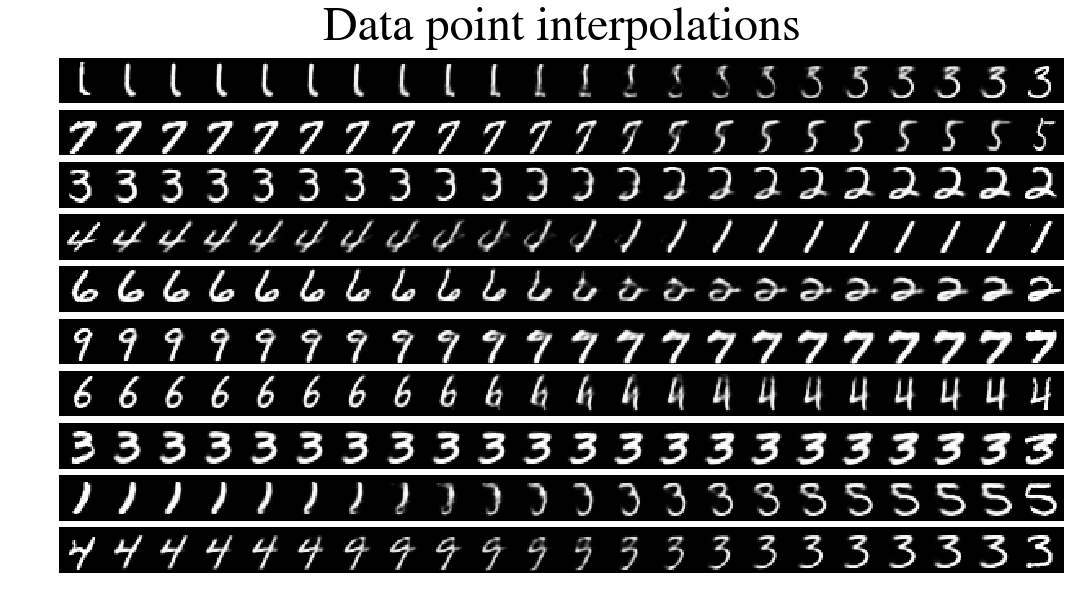}
    \caption{}
    \label{fig:point_inter}
\end{subfigure}
\begin{subfigure}{0.49\textwidth}
    \centering\includegraphics[width=\textwidth]{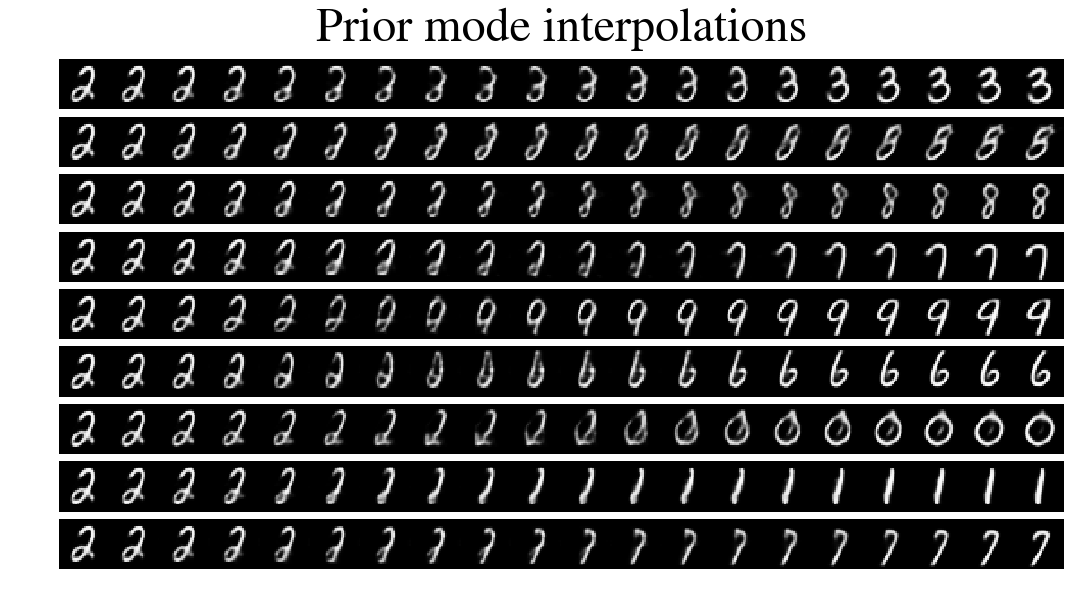}
    \caption{}
    \label{fig:prior_inter}
\end{subfigure}
\caption{Reconstructions from linear interpolations in the continuous latent space between two data points (a), and between the prior mode $\mu_2^0$, and the other nine prior modes $\mu_{k\neq2}^0$ (b). In (a), the true data points are shown in the first and last column next to their direct reconstructions.}
\label{fig:interpolations}
\end{center}
\end{figure}

GM-WAE has a smooth manifold structure in its latent variables. In Figure~\ref{fig:point_inter} the reconstructions of a linear interpolation with uniform step size in the continuous latent space is shown between pairs of data points. This compares similarly to other WAE and VAE approaches to MNIST. 
In Figure~\ref{fig:prior_inter} a linear interpolation is performed between the prior mode $\mu_2^0$, and the other nine prior modes $\mu_{k\neq2}^0$. This not only shows the smoothness of the learned latent manifold in all directions around a single mode of the prior, but also shows that the variatonal distribution has learned to match the modes of the prior. As one would hope given the suitability of a 10-mode GM-LVM to MNIST, almost every mode of the prior now represents a different digit. This level of control built into the prior requires not only a multi-modal prior, but also a training procedure that actually leverages the structure in both the prior and variational distribution, which seems to not be the case for VAEs (see Section~\ref{sec:VAE_difficulty}).

The quality of samples from our GM-WAE is related to the ability of the encoder networks to match the prior distribution. Figure~\ref{fig:generation_tight} and \ref{fig:prior_inter} demonstrate that the latent manifold learned is similar to the prior. Near the modes of the prior the samples are credible handwritten digits, with the encoder networks able to capture the structure within each mode of the data manifold (variation within each column) and clearly separate each different mode (variation between rows).

\begin{figure}
\begin{subfigure}{0.32\textwidth}
    \centering\includegraphics[width=\textwidth]{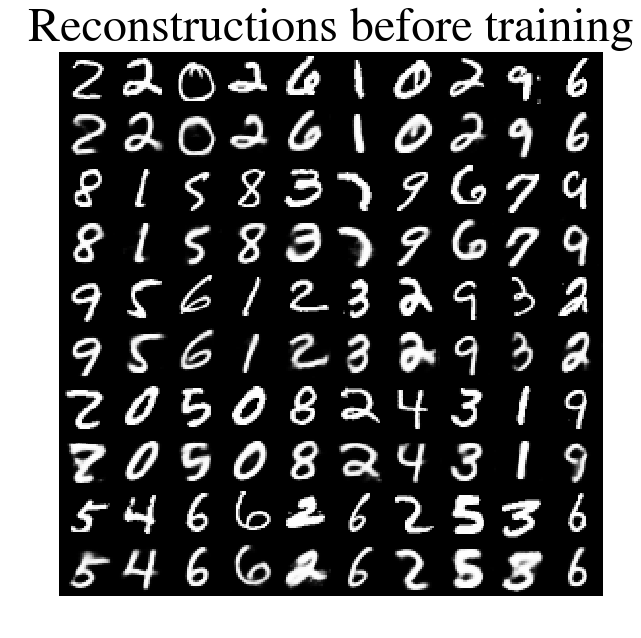}
    \caption{}
    \label{fig:recon_WAE_init_0}
\end{subfigure}
\hspace{.01\textwidth}
\begin{subfigure}{0.31\textwidth}
    \centering\includegraphics[width=\textwidth]{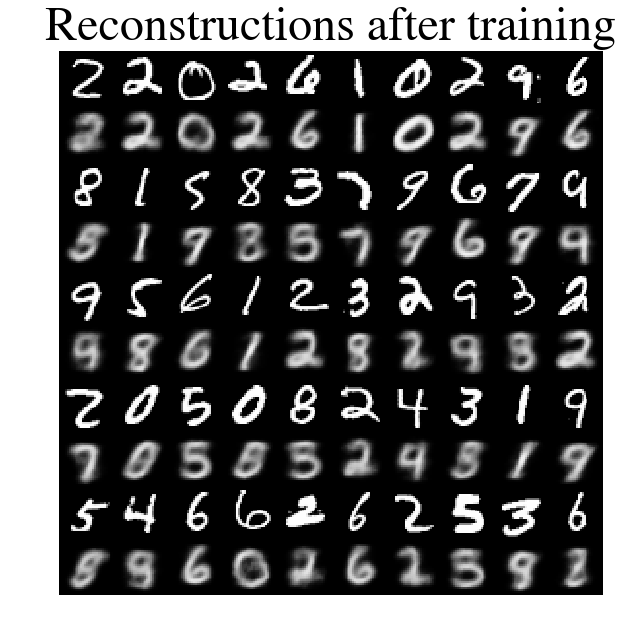}
    \caption{}
    \label{fig:recon_WAE_init_299}    
\end{subfigure}
\hspace{.01\textwidth}
\begin{subfigure}{0.32\textwidth}
    \centering\includegraphics[width=\textwidth]{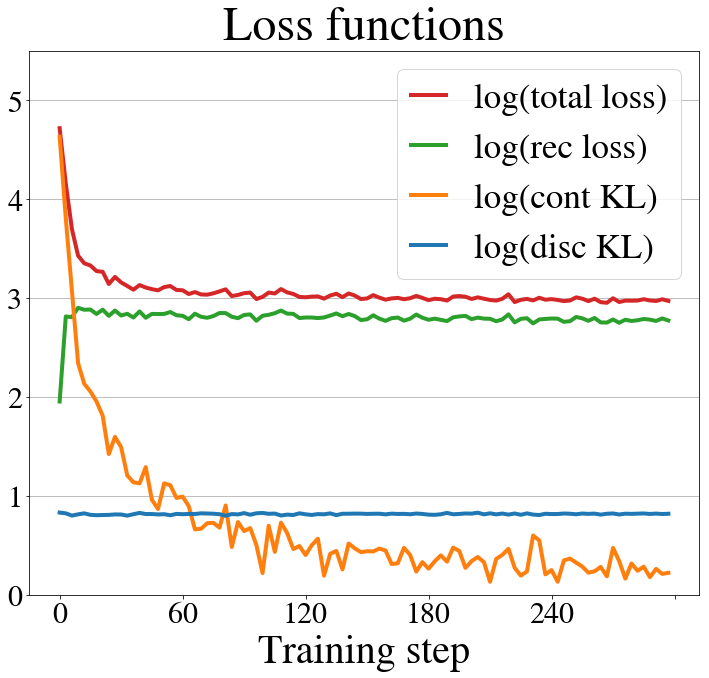}
    \caption{}
    \label{fig:loss_WAE_init}    
\end{subfigure}
\caption{(a) Reconstructions from an untrained VAE initialized with same parameters as our trained WAE. (b) Those same reconstructions after a few hundred training steps according to the VAE objective. (c) Learning curves for this training.}
\label{fig:WAE_init_VAE}
\end{figure}

We have argued that the VAE objective itself was responsible for the collapse of certain $k$ values in the discrete variational distribution, and that this was due to the per-data-point nature of the KL divergence in the VAE objective. To test this hypothesis, and to compare directly our trained WAE with the equivalent VAE that was discussed in Section~\ref{sec:VAE_difficulty}, we initialize the VAE with the parameters of the final trained WAE and train it according to the VAE objective. At initialization, the VAE with trained WAE parameters produces high quality samples and reconstructions (see Figure~\ref{fig:recon_WAE_init_0}). However, 
after a few hundred iterations, the reconstructions deteriorate significantly as shown in Figure~\ref{fig:recon_WAE_init_299} (further training does not improve these). The learning curves over the period of training between Figure~\ref{fig:recon_WAE_init_0} and \ref{fig:recon_WAE_init_299} are shown in Figure~\ref{fig:loss_WAE_init}, where the cause of the performance deterioration is clear: the continuous KL term in the VAE objective is multiple orders of magnitude larger than the reconstruction term, causing optimization to sacrifice reconstruction to reduce this KL term. Of course, the approximate posterior aggregated over the data will not be far from the prior as that distance is minimized in the WAE objective. However, this is not enough to ensure the VAE KL divergence term (for the continuous latent) is small, resulting in poor performance by the VAE as compared to the WAE.

Overall, the results for GM-WAE are qualitatively competitive with other approaches \cite{WAE}, despite a relatively low-complexity implementation. Moreover, given the improved latent-variable structure of the model, such a generative model provides more control over generation and inference.

\subsection{Latent variable fidelity}
\label{sec:LV_control}

In the previous section, it was shown that GM-WAE is able to both reconstruct data and generate new samples meaningfully from the prior distribution. We now turn to studying the variational distributions directly, including with how much fidelity a given class of digits is paired with a given discrete latent.

\begin{figure}
\begin{center}
\begin{subfigure}{0.42\textwidth}
    \centering\includegraphics[width=\textwidth]{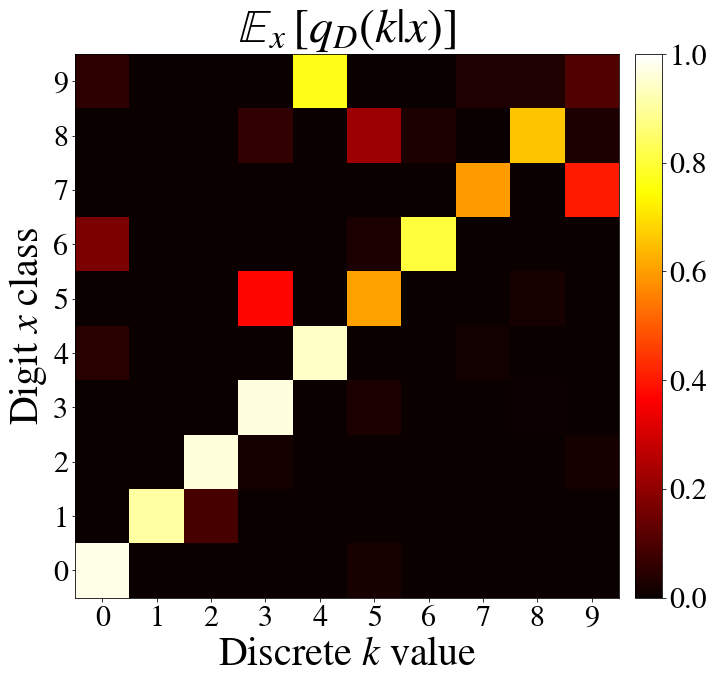}
    \caption{}
    \label{fig:qk_probs}
\end{subfigure}
\hspace{.05\textwidth}
\begin{subfigure}{0.45\textwidth}
    \centering\includegraphics[width=\textwidth]{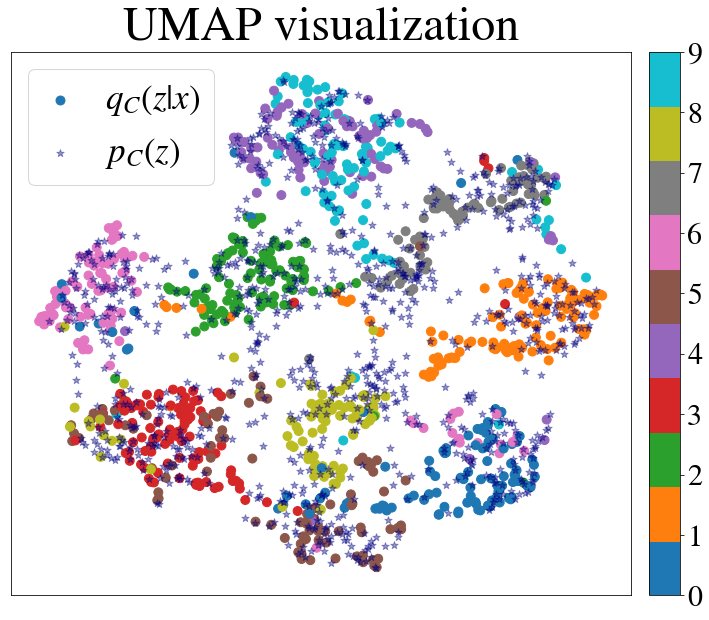}
    \caption{}
    \label{fig:qz_probs}
\end{subfigure}
\caption{Visualization of the variational distributions. (a) shows $\mathbb{E}_{\{x|\text{label}(x)=\ell\}} q_D(k|x)$ in row $\ell$. (b) shows $z|x\sim \sum_k q_C(z|k,x)q_D(k|x)$ dimensionally reduced using UMAP \cite{2018arXivUMAP}. 1000 encoded test-set digits and 1000 samples from the prior are used. Samples a colored by the digit label.}
\label{fig:q_distros}
\end{center}
\end{figure}

Consider first the discrete distribution $q_D(k|x)$ shown in Figure~\ref{fig:qk_probs}, where $\mathbb{E}_{\{x|\text{label}(x)=\ell\}} q_D(k|x)$ is shown in row $\ell$. From the staircase structure, it is clear that this distribution learns to approximately assign each discrete latent value $k$ to a different class of digit. However, it does not do so perfectly. This is expected as the GM-WAE seeks only to reconstruct the data from its encoding, not to encode it in any particular way. However, this does not mean GM-WAE is failing to use its discrete latent effectively. Indeed, when comparing Figure~\ref{fig:generation_tight} and Figure~\ref{fig:qk_probs}, a meaningful source of overlap between different values of $k$ and a single digit class can be seen. For example, in Figure~\ref{fig:qk_probs} the digit 9 is assigned partially to $k=0$, $k=4$, and $k=9$. In Figure~\ref{fig:generation_tight}, 9s drawn with a big-round loop are similar to digit 0, 9s with a small loop and long tail are similar to digit 7, and 9s with a balanced-size loop and tail are similar to digit 4. A similar discussion holds for 3s and 5s as well.

To assess the digit-class fidelity of the discrete encoder more quantitatively, we calculate the accuracy of the digit-class assignment according to $q_D(k|x)$. To assign a digit-class label to each $k$ value, we follow  a similar protocol to that of \cite{AAE}: we assign the digit-class label to the $k$ value that maximizes the average discrete latent for that class, in decreasing order of that maximum. With this simple assignment protocol, our GM-WAE achieves an accuracy on the held-out test set of $76\%$. For reference, basic $K$-means clustering \cite{kmeans} achieves $50$-$60\%$, and \cite{AAE} achieve $90\%$ (using $16$ discrete classes, and substantially different model and training procedure).

Another way to study the latent variable structure of GM-WAE is to consider dimensionally reduced visualizations of the continuous latent $z$. In Figure~\ref{fig:qz_probs} such a visualization is shown using UMAP \cite{2018arXivUMAP}. Distinct clusters can indeed be seen in the prior and in the samples from $q_C(z|k,x)$. Though the clusters of $z\sim q_C(z|k,x)$ do not fully align with those from the prior $z\sim p_D(z|k)$, they maintain significant overlap. The samples from $q_C(z|k,x)$ in Figure~\ref{fig:qz_probs} are colored according to the true digit labels, and show how GM-WAE learned to assign the digits to the different clusters. In particular, the 4 / 9 cluster is clearly overlapping, as seen in Figures~\ref{fig:qk_probs}, \ref{fig:generation_unfettered} and  \ref{fig:generation_tight}.

In this section we have see that the GM-WAE model is highly suited to the problem under study. It reconstructs data and provides meaningful samples, it effectively uses both discrete and continuous variational distributions, all while maintaining close proximity between the variational distribution and the prior.

\section{Conclusions}

We have studied an unsupervised generative model with a mixture of Gaussians latent variable structure, very well suited for data sets containing discrete classes of objects with continuous variation within each class. We discussed why such models are difficult to train in the Variational Autoencoder framework, and showed that a natural framework for training such models is given by Optimal Transport, in particular the Wasserstein Autoencoder. We found promising results training our model on MNIST, and demonstrated the additional control available to a highly structured model with both discrete and continuous latent variables. We hope this motivates further study of the exciting but nascent field of Optimal Transport in generative modeling.

\section{Acknowledgments}

This work was supported by the Alan Turing Institute under the EPSRC grant EP/N510129/1 and by AWS Cloud Credits for Research. The authors thank Aleksander Botev and Giulia Luise for helpful discussions on Optimal Transport, as well as Aleksander Botev, Roberto Fierimonte, Alex Mansbridge and Hippolyt Ritter for comments on various drafts of this paper.

\bibliography{Wasserstein}
\bibliographystyle{plain}

\end{document}

%% file: tikzstuff.tex
\usepackage{tikz}
\usetikzlibrary{arrows,shapes,backgrounds,through,shadows}

\tikzstyle{cont}=[circle, draw,thick,minimum size=7.5mm,line width=1pt,>=stealth]  

\tikzstyle{obs}=[fill=blue!10,thick]  

\tikzstyle{contobs}+=[cont]
\tikzstyle{contobs}+=[obs]
\tikzstyle{discobs}+=[disc]
\tikzstyle{discobs}+=[obs]

\tikzstyle{dgraph}=[->, line width=1.5pt]

%% file: NIPS_arXiv_version.bbl
\begin{thebibliography}{10}

\bibitem{ArjovskyB17}
M.~Arjovsky and L.~Bottou.
\newblock Towards principled methods for training generative adversarial
  networks.
\newblock In {\em International Conference on Learning Representations}, 2017.

\bibitem{WGAN}
M.~Arjovsky, S.~Chintala, and L.~Bottou.
\newblock {W}asserstein generative adversarial networks.
\newblock In {\em International Conference on Machine Learning}, 2017.

\bibitem{Bousquetetal17}
O.~Bousquet, S.~Gelly, I.~Tolstikhin, C.~J. Simon-Gabriel, and
  B.~Sch{\"o}lkopf.
\newblock From optimal transport to generative modeling: the {VEGAN} cookbook.
\newblock Technical report, 2017.

\bibitem{brooks2011handbook}
S.~Brooks, A.~Gelman, G.~Jones, and M.~Xiao-Li.
\newblock {\em Handbook of {M}arkov chain monte carlo}.
\newblock CRC press, 2011.

\bibitem{HackGMVAE}
N.~Dilokthanakul, P.~A.~M. Mediano, M.~Garnelo, M.~C.~H. Lee, H.~Salimbeni,
  K.~Arulkumaran, and M.~Shanahan.
\newblock Deep unsupervised clustering with {G}aussian mixture variational
  autoencoders.
\newblock {\em arXiv/1611.02648}, 2016.

\bibitem{gen_MMD}
G.~K. Dziugaite, D.~M. Roy, and Z.~Ghahramani.
\newblock Training generative neural networks via maximum mean discrepancy
  optimization.
\newblock In {\em Conference on Uncertainty in Artificial Intelligence}, 2015.

\bibitem{FastInferRepeat}
S.~M.~A. Eslami, N.~Heess, T.~Weber, Y.~Tassa, D.~Szepesvari, K.~Kavukcuoglu,
  and G.~E. Hinton.
\newblock Attend, infer, repeat: fast scene understanding with generative
  models.
\newblock In {\em Advances in Neural Information Processing Systems}, 2016.

\bibitem{GANs}
I.~Goodfellow, J.~Pouget-Abadie, M.~Mirza, B.~Xu, D.~Warde-Farley, S.~Ozair,
  A.~Courville, and Y.~Bengio.
\newblock Generative adversarial networks.
\newblock In {\em Advances in Neural Information Processing Systems}. 2014.

\bibitem{MMD}
A.~Gretton, K.~M. Borgwardt, M.~J. Rasch, B.~Sch\"{o}lkopf, and A.~Smola.
\newblock A kernel two-sample test.
\newblock {\em Journal of Machine Learning Research}, 2012.

\bibitem{kernel_choice}
A.~Gretton, D.~Sejdinovic, H.~Strathmann, S.~Balakrishnan, M.~Pontil,
  K.~Fukumizu, and B.~K. Sriperumbudur.
\newblock Optimal kernel choice for large-scale two-sample tests.
\newblock In {\em Advances in Neural Information Processing Systems}, 2012.

\bibitem{ELBO_surgery}
M.~D. Hoffman and M.~J. Johnson.
\newblock {ELBO} surgery: yet another way to carve up the variational evidence
  lower bound.
\newblock In {\em NIPS Workshop on Advances in Approximate Bayesian Inference},
  2016.

\bibitem{BatchNorm}
S.~Ioffe and C.~Szegedy.
\newblock Batch normalization: accelerating deep network training by reducing
  internal covariate shift.
\newblock In {\em International Conference on Machine Learning}, 2015.

\bibitem{GumbelSoftmax}
E.~Jang, S.~Gu, and B.~Poole.
\newblock Categorical reparameterization with gumbel-softmax.
\newblock In {\em International Conference on Learning Representations}, 2017.

\bibitem{SVAE}
M.~J. Johnson, D.~Duvenaud, A.~B. Wiltschko, S.~R. Datta, and R.~P. Adams.
\newblock Composing graphical models with neural networks for structured
  representations and fast inference.
\newblock In {\em Advances in Neural Information Processing Systems}. 2016.

\bibitem{ADAM}
D.~P. Kingma and J.~Ba.
\newblock Adam: a method for stochastic optimization.
\newblock In {\em International Conference on Learning Representations}, 2015.

\bibitem{kingma2014semi}
D.~P. Kingma, S.~Mohamed, D.~J. Rezende, and M.~Welling.
\newblock Semi-supervised learning with deep generative models.
\newblock In {\em Advances in Neural Information Processing Systems}, 2014.

\bibitem{kingma2014stochastic}
D.~P. Kingma and M.~Welling.
\newblock Auto-encoding variational {B}ayes.
\newblock In {\em International Conference on Learning Representations}, 2014.

\bibitem{Lawson2017LearningHA}
D.~Lawson, G.~Tucker, C.-C. Chiu, C.~Raffel, K.~Swersky, and N.~Jaitly.
\newblock Learning hard alignments with variational inference.
\newblock In {\em IEEE International Conference on Acoustics, Speech and Signal
  Processing}, 2018.

\bibitem{moments_matching}
Y.~Li, K.~Swersky, and R.~Zemel.
\newblock Generative moment matching networks.
\newblock In {\em International Conference on Machine Learning}, 2015.

\bibitem{kmeans}
J.~MacQueen.
\newblock Some methods for classification and analysis of multivariate
  observations.
\newblock In {\em Proceedings of the Fifth Berkeley Symposium on Mathematical
  Statistics and Probability, Volume 1: Statistics}, Berkeley, Calif., 1967.
  University of California Press.

\bibitem{Concrete}
C.~J. Maddison, A.~Mnih, and Y.~W. Teh.
\newblock The concrete distribution: a continuous relaxation of discrete random
  variables.
\newblock In {\em International Conference on Learning Representations}, 2017.

\bibitem{AAE}
A.~Makhzani, J.~Shlens, N.~Jaitly, and I.~Goodfellow.
\newblock Adversarial autoencoders.
\newblock In {\em International Conference on Learning Representations}, 2016.

\bibitem{2018arXivUMAP}
L.~{McInnes} and J.~{Healy}.
\newblock {UMAP}: uniform manifold approximation and projection for dimension
  reduction.
\newblock {\em arXiv/1802.03426}, 2018.

\bibitem{DCGAN}
A.~Radford, L.~Metz, and S.~Chintala.
\newblock Unsupervised representation learning with deep convolutional
  generative adversarial networks.
\newblock In {\em International Conference on Learning Representations}, 2015.

\bibitem{rezende2014stochastic}
D.~J. Rezende, S.~Mohamed, and D.~Wierstra.
\newblock Stochastic backpropagation and approximate inference in deep
  generative models.
\newblock In {\em International Conference on Machine Learning}, 2014.

\bibitem{WAE_followup}
P.~K. Rubenstein, B.~Schoelkopf, and I.~Tolstikhin.
\newblock On the latent space of {W}asserstein auto-encoders.
\newblock In {\em Workshop track - International Conference on Learning
  Representations}, 2018.

\bibitem{WAE}
I.~Tolstikhin, O.~Bousquet, S.~Gelly, and B.~Schoelkopf.
\newblock {W}asserstein auto-encoders.
\newblock In {\em International Conference on Learning Representations}, 2018.

\bibitem{NeuralDiscrete}
A.~Van~den Oord, O.~Vinyals, and K.~kavukcuoglu.
\newblock Neural discrete representation learning.
\newblock In {\em Advances in Neural Information Processing Systems}, 2017.

\bibitem{villani2008optimal}
C.~Villani.
\newblock {\em Optimal Transport: Old and New}.
\newblock Grundlehren der mathematischen Wissenschaften. Springer Berlin
  Heidelberg, 2008.

\end{thebibliography}
